\documentclass[conference, 9pt]{IEEEtran}
\IEEEoverridecommandlockouts
\usepackage{cite}
\usepackage{amsmath,amssymb,amsfonts}
\usepackage{algorithmic}
\usepackage{graphicx}
\usepackage{textcomp}
\usepackage{xcolor}
\usepackage{multirow}
\usepackage{array} 
\usepackage{makecell} 
\usepackage{booktabs} 
\usepackage{url}
\usepackage[breaklinks=true,letterpaper=true,colorlinks,bookmarks=false]{hyperref}

\def\BibTeX{{\rm B\kern-.05em{\sc i\kern-.025em b}\kern-.08em
    T\kern-.1667em\lower.7ex\hbox{E}\kern-.125emX}}
\begin{document}

\title{TA-V2A: Textually Assisted Video-to-Audio Generation\\
\thanks{This work is supported by the National Key Research and Development Program of China (No.2024YFB2808902), and the High-performance Computing Platform of Peking University.}
}
\author{
\IEEEauthorblockN{
\begin{minipage}[t]{0.3\textwidth}
\centering
Yuhuan You\\
\textit{State Key Laboratory of\\ General Artificial Intelligence}\\
\textit{School of \\ Intelligence Science and Technology}\\
\textit{Peking University}, 
Beijing, China \\
2000017809@stu.pku.edu.cn
\end{minipage}
\hfill
\begin{minipage}[t]{0.3\textwidth}
\centering
Xihong Wu\\
\textit{State Key Laboratory of\\ General Artificial Intelligence}\\
\textit{School of \\ Intelligence Science and Technology}\\
\textit{Peking University}, 
Beijing, China \\
wxh@cis.pku.edu.cn
\end{minipage}
\hfill
\begin{minipage}[t]{0.3\textwidth}
\centering
Tianshu Qu\\
\textit{State Key Laboratory of\\ General Artificial Intelligence}\\
\textit{School of \\ Intelligence Science and Technology}\\
\textit{Peking University}, 
Beijing, China \\
qutianshu@pku.edu.cn
\end{minipage}
}
}

\maketitle
\begin{abstract}
As artificial intelligence-generated content (AIGC) continues to evolve, video-to-audio (V2A) generation has emerged as a key area with promising applications in multimedia editing, augmented reality, and automated content creation. While Transformer and Diffusion models have advanced audio generation, a significant challenge persists in extracting precise semantic information from videos, as current models often lose sequential context by relying solely on frame-based features. To address this, we present TA-V2A, a method that integrates language, audio, and video features to improve semantic representation in latent space. By incorporating large language models for enhanced video comprehension, our approach leverages text guidance to enrich semantic expression. Our diffusion model-based system utilizes automated text modulation to enhance inference quality and efficiency, providing personalized control through text-guided interfaces. This integration enhances semantic expression while ensuring temporal alignment, leading to more accurate and coherent video-to-audio generation.
\end{abstract}

\section{Introduction}

Recently, the specific modality transformation task of video-to-audio has garnered considerable attention. The ability to generate corresponding audio from video is crucial for applications such as enhancing virtual reality experiences, automated video foley synthesis, and improving the performance of robots in perceiving and understanding environments.

V2A tasks generally derive from text-to-audio (T2A) frameworks, such as \cite{liu2023audioldm, kreuk2022audiogen}, which focus solely on text input for audio generation, while some approaches, including \cite{mo2024t2av, wang2024v2a}, extend T2A frameworks by treating video frames as auxiliary features to enhance the audio generation process.
Other approaches, such as \cite{yang2023cmmd, kurmi2021collaborative, ruan2023mm}, aim for joint audio and video generation. Methods employing Generative Adversarial Networks (GANs) \cite{ghose2022foleygan, iashin2021SpecVQGAN} and Transformers \cite{du2023conditional, pascual2024masked} have also been utilized for pure V2A tasks. Recent advances in diffusion-based technologies have further expanded the capabilities of audio generation, with Luo et al.\cite{luo2024diff} leveraging Latent Diffusion Models (LDM) and Contrastive Audio-Video Pretraining (CAVP) for V2A tasks and Xu et al.\cite{xu2024vta-ldm} delving into the impact of module selection within the V2A framework.

Despite these advancements, a key challenge still remains in extracting precise semantic information from videos, as current models often lose sequential context when relying solely on frame-based features. This lack of temporal consistency in many models has led to gaps in audio generation, where the produced sound fails to match the nuances of actions or events unfolding over time. To address this, integrating text guidance has shown promise in enhancing semantic representation. Previous work has utilized text to guide audio generation \cite{jeong2024read} and combined video and text features for better semantic and temporal alignment \cite{zhang2024foleycrafter}.

Building on these efforts, we introduce TA-V2A, a text-assisted V2A generation system, which further leverages text as an auxiliary feature to participate in network training, feature generation, and the guidance stages of the diffusion process. The role of text assistance in this system is twofold: first, we integrate video, audio, and text modalities in a unified training framework to achieve precise and refined semantic alignment; second, inspired by advancements in multimodal large language models (MLLMs) \cite{wu2023next, tang2024any, damonlpsg2024videollama2}, we employ large language models to generate textual descriptions of videos, which are then used for text-aligned training and as guiding conditions within the diffusion model. This approach significantly enhances the generation of audio that closely aligns with human descriptive preferences.

Simultaneously, we delve into the exploration of feature vectors within the latent space that are most conducive to high-quality audio generation in V2A tasks, as well as the decomposition and integration of multi-condition inputs in the diffusion model. By focusing on the latent space, we aim to better understand the intricate relationships between video, audio, and text features, ensuring that the generated audio not only aligns with the video's temporal sequence but also accurately reflects the semantic content captured in the video. 
Extensive experiments and analyses validate the effectiveness of these approaches, suggesting that TA-V2A can significantly advance video-to-audio generation, enhancing multimedia processing and intelligent information comprehension.

The rest of this paper is organized as follows. Section \ref{sec:method} presents our proposed approach and introduces the core methodologies . Section \ref{sec:exp} details the experimental configurations and results. Finally, Section \ref{sec:con} offers conclusions.

\section{Method}
\label{sec:method}

\begin{figure*}[tb]
      \centering
      \centerline{
        \includegraphics[width=\textwidth]{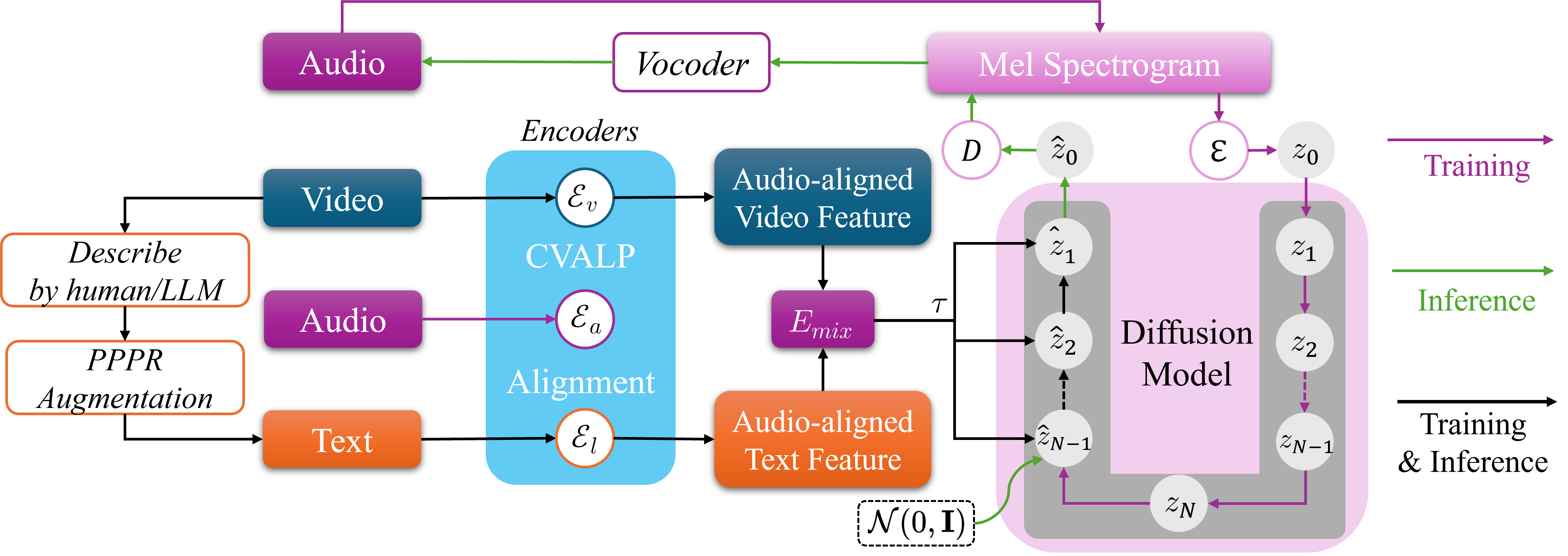}
        }

\caption{The complete workflow of the TA-V2A generation system. The system takes video and textual descriptions as input, with the textual description generated by an LLM. The CVALP module extracts and aligns features from video, audio, and text, creating audio-aligned video and text features. These features are then fed into LDM, which iteratively generates high-quality audio from noise. During inference, guidance techniques such as CFG and human-modified text prompts are used to control the generation process, ensuring better alignment between the generated audio and the input modalities. Finally, the audio representation is decoded into a Mel-spectrogram and synthesized into the actual audio waveform using a vocoder.}
    \label{fig:frame}
\end{figure*}

Our TA-V2A generation system combines video, audio, and textual data to produce synchronized audio outputs from video inputs using contrastive learning, feature alignment, and generative modeling. An overview of the complete workflow is illustrated in Fig. \ref{fig:frame}.

The system begins with video and textual inputs, where text is either manually provided or generated by an LLM and undergoes automated augmentation. Features from video, audio, and text are extracted by modality-specific encoders \( \mathcal{E}_v, \mathcal{E}_a, \mathcal{E}_l \) in the Contrastive Video-Audio-Language Pretraining (CVALP) module, aligned, and fused into audio-aligned features \( E_\text{mix} \). These features are fed into LDM, which generates audio features from Gaussian noise \( \mathcal{N}(0, I) \).
The LDM output is decoded into a Mel-spectrogram (\( \hat{z}_0 \)) and synthesized into an audio waveform by a vocoder. Training refines system parameters, while inference generates the final synchronized audio output.
Below, we detail each component of the system.

\subsection{Contrastive Video-Audio-Language Pretraining}
\label{ssec:pretrain}

The CVALP module serves as the backbone of our feature extraction and alignment process. Inspired by prior works \cite{radford2021clip, wu2023clap, luo2024diff}, CVALP aligns video and text modalities with audio using contrastive learning across video-audio and text-audio pairs. This simultaneous alignment improves feature quality, convergence speed, model robustness, and hierarchical learning.

Given a video-audio-text triplet \((x_v, x_a, x_l)\), where \(x_v \in \mathbb{R}^{T_v \times 3 \times H \times W}\) denotes a video clip with \(T_v\) frames of size \(H \times W\) and RGB channels, \(x_a \in \mathbb{R}^{T_a \times M}\) represents a Mel-spectrogram with \(T_a\) time steps and \(M\) mel bands, and \(x_l\) is the corresponding text description. We use video, audio, and text encoders \(f_V(\cdot)\), \(f_A(\cdot)\), and \(f_L(\cdot)\) to extract features \(E_v^T, E_a^T \in \mathbb{R}^{T \times C}\) from video and audio, where \(T = 32\) is the number of temporal segments and \(C = 512\) is the feature dimension, and \(E_l \in \mathbb{R}^C\) from text. To align feature dimensions, we apply temporal pooling to the video and audio features, resulting in \(E_v\) and \(E_a \in \mathbb{R}^C\). This ensures all features are in the same space \(\mathbb{R}^C\), facilitating fusion and comparison.

We define the contrastive loss function for the \(i\)-th cross-modal pair as:

\begin{equation}
    f(E_x^i, E_y^i, M) = \log \frac{\exp(E_x^i \cdot E_y^i / \tau)}{\sum_{j=1}^{M} \exp(E_x^i \cdot E_y^j / \tau)} 
\end{equation}
where \(E_x^i\) and \(E_y^i\) are feature vectors from different modalities \(x\) and \(y\), \(M\) is the number of cross-modal pairs, and \(\tau\) controls softmax smoothness. The numerator represents the similarity between the correct pair, while the denominator sums the similarities between \(E_x^i\) and all \(E_y^j\) pairs. This loss is not permutation symmetric because swapping \(E_x^i\) and \(E_y^i\) changes the calculation, as \(E_x^i\) is always the anchor point compared to all \(E_y^j\). This asymmetry is useful when one modality, such as text, provides stronger semantic guidance for aligning features from another modality, such as video or audio.

For the feature triplet $(E_v, E_a, E_l)$, we define the loss function centered on audio as follows:
\begin{equation}
    \mathcal{L}_{AL}^i = f(E_a^i, E_l^i, N) + f(E_l^i, E_a^i, N)
\end{equation}
\begin{equation}
    \mathcal{L}_{AVS}^i = f(E_a^i, E_v^i, N_S) + f(E_v^i, E_a^i, N_S)
\end{equation}
\begin{equation}
    \mathcal{L}_{AVT}^i = f(E_a^i, E_v^i, N_T) + f(E_v^i, E_a^i, N_T)
\end{equation}
\begin{equation}
    \mathcal{L} = \frac{1}{2N}\sum_{i=1}^N \mathcal{L}_{AL}^i
    +
    \frac{\lambda}{2N_S}\sum_{i=1}^{N_S} \mathcal{L}_{AVS}^{i}
    +
    \frac{\mu}{2N_T}\sum_{i=1}^{N_T} \mathcal{L}_{AVT}^{i}
\end{equation}
where $E_a^i$, $E_l^i$, and $E_v^i$ represent the audio, text, and video embeddings for the $i$-th sample, respectively. \(N\), \(N_S\), and \(N_T\) represent the total audio-text pairs, cross-video audio-video pairs, and intra-video temporal pairs, as defined in \cite{luo2024diff}.

The parameters $\lambda$ and $\mu$ are weights that determine the relative importance of the different loss components: $1/(\lambda+\mu)$ reflects the importance of text relative to video, and $\mu/(1+\lambda)$ indicates the emphasis on temporal alignment versus semantic expression. They will be decided by experiments.

Equations (2) capture the similarity measure between audio and text, while (3) and (4) express the similarity between audio and video from semantic and temporal perspectives. The final loss function in (5) integrates these modality pairs.

\subsection{Feature Mixing}
\label{ssec:mix}

To blend the features extracted from different modalities, we employ feature mixing strategies that balance information from video and text.
Since video modality contains both semantic and temporal features, while text usually only contains semantic features, blindly aligning the two (such as using methods like cross-attention) may likely lead to the loss of temporal alignment information learned in CAVP. To address this, we employ weighted averaging and concatenation methods for feature fusion:
\begin{equation}
    E_\text{mix}^\text{aver} = E_\text{aver} = 
    \left[E_l + (\lambda + \mu) E_v\right]/(1 + \lambda + \mu) 
\end{equation}
\begin{equation}  
        E_\text{concat} = 
    \text{concat}[P(E_v); P(E_l)] \}
\end{equation}
where \(P(E_v)\) and \(P(E_l)\) are linear projections reducing dimensions by half to keep \(E_\text{concat}\) the same size.

We utilize positional encoding and a projection layer \(\tau_{\theta}\) to map \(E_\text{concat}\) to the appropriate dimension, expressed as \({E}_\text{mix}^\text{concat} = \tau_{\theta}(E_\text{concat}) = MLP(E_\text{concat} + PE)\), where MLP serves as the projection layer and PE denotes positional encoding. This hybrid feature representation is used as an input feature vector for training and inference in LDM.
\subsection{Latent Diffusion Model}
\label{ssec:LDM}

LDM generates high-dimensional audio data in a lower-dimensional latent space\cite{rombach2022ldm}. Starting from an encoded Mel-spectrogram $z_0 = \mathcal{E}(x_a)$, the diffusion process is conducted in the latent space, where noise is progressively added to $z_0$, forming a sequence of latent variables $z_1, z_2, \ldots, z_T$. Each diffusion step is modeled as:
\begin{equation}
    q(z_t | z_{t-1}) = \mathcal{N}(z_t; \sqrt{1 - \beta_t} z_{t-1}, \beta_t \mathbf{I})
\end{equation}
where $t$ denotes the time step, $\beta_t$ is the diffusion coefficient, and $\mathcal{N}$ represents the normal distribution. For data generation, the model reverses the diffusion process, starting from a sample $z_T$ drawn from a Gaussian distribution. A neural network $p_\theta$ predicts the reverse step at each timestep:
\begin{equation}
    p_\theta(z_{t-1} | z_t) = \mathcal{N}\left(z_{t-1}; \mu_\theta(z_t, t, E_\text{mix}), \sigma_t^2\mathbf{I} \right)
\end{equation}
where \(\mu_\theta\) represents the predicted mean of the Gaussian distribution by the neural network parameterized by \(\theta\), and \(\sigma_t\) denotes the standard deviation, often related to the timestep \(t\).

After the reverse diffusion, the final latent variable $z_0$ is decoded by $\mathcal{D}$ back into the data space, producing the generated Mel-Spectrogram $\hat{x}_a = \mathcal{D}(z_0)$, which is then converted into an audio sample using a vocoder. The conditional loss function is given by:
\begin{equation}
    \mathcal{L}_\text{LDM} = \mathbb{E}_{\mathcal{E}(x), \epsilon \sim \mathcal{N}(0,1), t}\left[\left\|\epsilon - \epsilon_\theta\left(z_t, t, E_\text{mix}\right)\right\|_2^2\right]
\end{equation}

\subsection{Inference with Guidance}
\label{ssec:guidance}

During inference, guidance techniques such as Classifier Guidance (CG)\cite{dhariwal2021cg} and Classifier-Free Guidance (CFG)\cite{ho2022cfg} are employed to control the generation process. CG relies on an additional classifier $P_{\phi}$ to guide the reverse process at each timestep via the gradient of the class label log-likelihood $\nabla \log P_{\phi}(y|x_t)$. CFG, on the other hand, combines conditional and unconditional score estimates to steer the reverse process. As suggested in \cite{luo2024diff}, double guidance can be applied for enhanced alignment:
\begin{equation}
    \hat{\epsilon}_\theta\left(z_t, t\right) \leftarrow \begin{gathered}
    \omega \epsilon_\theta\left(z_t, t, E_{\text {mix}}\right)+(1-\omega) \epsilon_\theta\left(z_t, t, \varnothing \right) \\
    -\gamma \bar{\beta}_t \nabla \log P_\phi\left(y | z_t, t, E_v\right)
    \end{gathered}
\end{equation}
where $\gamma$ and $\omega$ represent the scales for CG and CFG, respectively. Notably, CFG employs $E_{\text{mix}}$, while CG uses $E_v$ due to the aligned classifier $P_{\phi}(y|z_t, E_v)$ trained for the alignment of audio-visual pairs as discussed in \cite{luo2024diff}.

From the perspective of Energy Based Models (EBMs)\cite{liu2022compositional}, multiple conditions can also influence the inference process independently without combination. The conditional probability can be estimated by the following formula:
\begin{equation}
    p\left(\boldsymbol{x} | \boldsymbol{c}_1, \ldots, \boldsymbol{c}_n\right) \propto p(\boldsymbol{x}) \prod_{i=1}^n \frac{p\left(\boldsymbol{x} | \boldsymbol{c}_i\right)}{p(\boldsymbol{x})}
\end{equation}

Here, we define \( g \) to represent the difference between the unconditional and conditional score estimates:
\begin{equation}
    g(c,\omega)=\omega\left(\epsilon_\theta\left(z_t, t | c\right)-\epsilon_\theta\left(z_t, t\right)\right)
\end{equation}
where \( c \) is the condition and \( \omega \) is a hyperparameter. Then the multi-conditioned inference steps could be represented as
\begin{equation}
    \hat{\epsilon}_\theta\left(z_t, t\right)
    \leftarrow 
    \begin{gathered}
        \epsilon_\theta\left(z_t, t\right)+g\left(E_l, \omega_l\right)-g\left(E_{n l}, \omega_{n l}\right)\\
        +g\left(E_v, \omega_v\right)
        -\gamma \bar{\beta}_t \nabla \log P_\phi\left(y | z_t, t, E_v\right)
    \end{gathered}
\end{equation}
where \( E_v \) represents video features, \( E_l \) represents prompt features, and \( E_{nl} \) represents negative prompt features. This approach allows for a more flexible inference process by independently considering the effects of multiple conditions.

The workflow is designed to efficiently process the video and textual descriptions through CVALP, align and mix features using LDM, and apply inference techniques to produce the final high-quality audio output that matches the given video content.
Speicfically, in the inference process, the positive prompt input by the human will undergo the Portable Plug-in Prompt Refiner (PPPR)\cite{shi2024pppr} for standardization, ensuring that it corresponds with the AI-generated text involved in the training. If a human-provided prompt is not available, we will use the Video-LlaMA2\cite{damonlpsg2024videollama2} model to generate a description of the video content automatically.

\section{Experiments}
\label{sec:exp}

\subsection{Datasets and Data Processing}
\label{ssec:data}

Our study utilizes the VGGSound \cite{chen2020vggsound} dataset, which contains approximately 200,000 videos, each with a duration of 10 seconds. As in the original scheme, we choose to use the provided training and testing split of the dataset, with 183,971 videos in the training set and 15,496 videos in the test set. Data preprocessing includes video, audio, and text processing: the videos were resized to $224 \times 224$ and frames were sampled at 4 FPS; the audio was sampled at 16kHz and converted to Mel spectrograms (Mel Basis $M = 128$), with a hop size uniformly set to 256.
For text processing, we expanded the annotations in the VGGSound dataset using ChatGPT4o to improve text-audio alignment and maintain consistency with inference prompts. A unified prompt was used to generate video descriptions, which served as contrastive learning material in CVALP training:

\textit{"Here are the annotated texts from a video dataset. Please expand each into a full sentence, keeping the core content unchanged."}

To introduce variation and avoid identical descriptions for similar videos during contrastive learning, we applied PPPR while preserving the core content, as detailed in \cite{shi2024pppr}.

\subsection{Configurations}
\label{ssec:config}

In the CVALP contrastive learning process, we employed the PANNs\cite{kong2020pannslargescalepretrainedaudio} model pretrained on the AudioSet\cite{gemmeke2017audioset} dataset as the audio encoder, and the SlowOnly\cite{feichtenhofer2019slowfastnetworksvideorecognition} model pretrained on the Kinetics-400\cite{kay2017kineticshumanactionvideo} dataset as the video encoder, with Flan-T5\cite{chung2022flant5} serving as the text encoder. For the LDM, we adopted the architecture of Stable Diffusion, utilizing frozen, pretrained latent encoder 
$\mathcal{E}$ and decoder 
$\mathcal{D}$ components. The denoising process involved 1,000 steps, and we used a learning rate of $10^{-4}$ with an initial warmup phase of 1,000 steps. During the inference stage, we incorporated agent attention\cite{han2024agentattentionintegrationsoftmax}, which introduces agent tokens to improve computational efficiency while maintaining global context modeling, to enhance speed and effectiveness.

We set the CFG scale to $\omega=4.5$ and the CG scale to $\gamma=50$ same as \cite{luo2024diff}. For composite condition inference, we set $\omega_l=\omega_{nl}=\omega_v=2.5$ by experiments. After generating the Mel-spectrogram, we used the GLA-GRAD \cite{liu2024gla} vocoder, which is also based on a diffusion model architecture, to produce the audio signal.

\subsection{Evaluation}
\label{ssec:eval}

\subsubsection{Evaluation Metrics}
\label{sssec:eval}

\quad

\noindent\textbf{Baseline}  
We conducted ablation experiments using the same LDM module, deriving latent features through different methods. Our model employed the CVALP module with averaging (aver.) and concatenation (concat) techniques. For comparison, we tested against the Diff-Foley \cite{luo2024diff} model with the CAVP module and the VTA-LDM \cite{xu2024vta-ldm} model using Clip4Clip \cite{luo2021clip4clip}. All experiments were conducted on the VGGSound \cite{chen2020vggsound} test set, generating 8-second audio clips for evaluation.

\noindent\textbf{Objective Evaluation}  
We used four metrics from \cite{Taming, xu2024vta-ldm} to assess semantic generation quality: Inception Score (IS) for evaluating how well the generated distribution matches the diversity of real data. Fréchet Inception Distance (FID) and Fréchet Audio Distance (FAD) are used to compare the statistical features of generated samples with those of real samples, as demonstrated by \cite{tailleur2024correlation}, which validates their effectiveness. Mean Kullback–Leibler divergence (MKL) to measure the divergence between the probability distributions of generated and real data, reflecting how closely they align. We also used Alignment Accuracy (Align) from \cite{luo2024diff} to evaluate temporal synchronization between generated audio and video content.

\noindent\textbf{Subjective Evaluation}  
Twenty participants (8 females, 12 males, aged 20-45), with with self-reported normal hearing and normal or corrected-to-normal vision, rated audio-visual clips using Sennheiser HD600 headphones.  Participants viewed five distinct videos, each with four generated audio tracks corresponding to the four experimental conditions listed in the table (as identified by the "Model" and "Latent Features" columns collectively), presented in random order. Each participant listened to each audio sample only once to ensure that the ratings reflected their initial impressions.
This procedure was repeated for each of the five videos. 

Among these conditions, the "Users" group contributed by writing descriptions based on the video content. These descriptions were used as text prompt input for the model in the "Users" condition, serving as a guiding condition during the diffusion process.

Participants rated the audio tracks using Mean Opinion Scores (MOS)\cite{ITUT_P800} on a five-point scale, where 1 represents "Bad," 2 "Poor," 3 "Fair," 4 "Good," and 5 "Excellent." MOS for semantic consistency was based on how well the audio content matched the video content, while temporal alignment MOS was based on the synchronization between the audio and the visual cues.

\subsubsection{Results and Analysis}
\label{sssec:result}
\begin{table}[tb]
\caption{Objective Evaluation Results}

    \begin{center}
        \begin{tabular}{c>{\centering\arraybackslash}m{1cm}>{\centering\arraybackslash}m{0.5cm}>{\centering\arraybackslash}m{0.5cm}>{\centering\arraybackslash}m{0.5cm}>{\centering\arraybackslash}m{0.5cm}>{\centering\arraybackslash}m{0.8cm}}
        \toprule
        Model & Features & IS↑ & FID↓ & FAD↓ & MKL↓ & Align(\%)↑ \\
        \midrule
        TA-V2A & \makecell{CVALP \\ (aver.)} & 7.16 & 48.47 & 6.03 & 4.92 & 72.61 \\
        TA-V2A & \makecell{CVALP \\ (concat)} & 10.59 & \textbf{21.71} & \textbf{2.66} & \textbf{2.74} & 84.37 \\
        Diff-Foley & CAVP & 9.51 & 36.20 & 4.87 & 4.53 & \textbf{86.77} \\
        VTA-LDM & Clip4Clip & \textbf{10.74} & 26.05 & 2.74 & 3.10 & 79.89 \\
        \bottomrule
        \end{tabular}
    \end{center}
    \label{tab:tab1}
\end{table}


\begin{table}[tb]
    \caption{Subjective Evaluation Results}
    \begin{center}
        \begin{tabular}{cccc}
        \toprule
        Model & Latent Features & Semantic↑ & Temporal↑ \\
        \midrule
        TA-V2A (Auto) & CVALP (concat) & 4.00 & \textbf{3.75} \\
        TA-V2A (Users) & CVALP (concat) & \textbf{4.30} & 3.70 \\
        Diff-Foley & CAVP & 3.35 & \textbf{3.75} \\
        VTA-LDM & Clip4Clip & 3.90 & 3.10 \\
        \bottomrule
        \end{tabular}
    \end{center}
    \label{tab:tab2}
\end{table}

\quad

To showcase our results more effectively, we provide a video that visually demonstrates the audio generation outcomes: \href{https://drive.google.com/file/d/1P3BLVhO_GcpYUqq67ij3LOD1FRxR3nra/view?usp=sharing}{Display}

\noindent \textbf{Objective Evaluation Results}
Table \ref{tab:tab1} presents the objective evaluation outcomes of the models. The TA-V2A model demonstrates a notably strong performance across all metrics when employing the concatenation (concat) method to derive CVALP features, achieving particularly outstanding results in the FID and FAD metrics. In contrast, the performance diminishes when utilizing the averaging (average) method, likely due to the simple averaging process disrupting the algebraic structure of the feature vectors or matrices. The approach of concatenating multimodal information, followed by projection and dimensionality reduction, proves effective in enhancing semantic expression while simultaneously preserving the alignment capabilities within the video features.
\begin{figure}[tb]
  \centering
  \centerline{
\includegraphics[width=0.95\columnwidth]{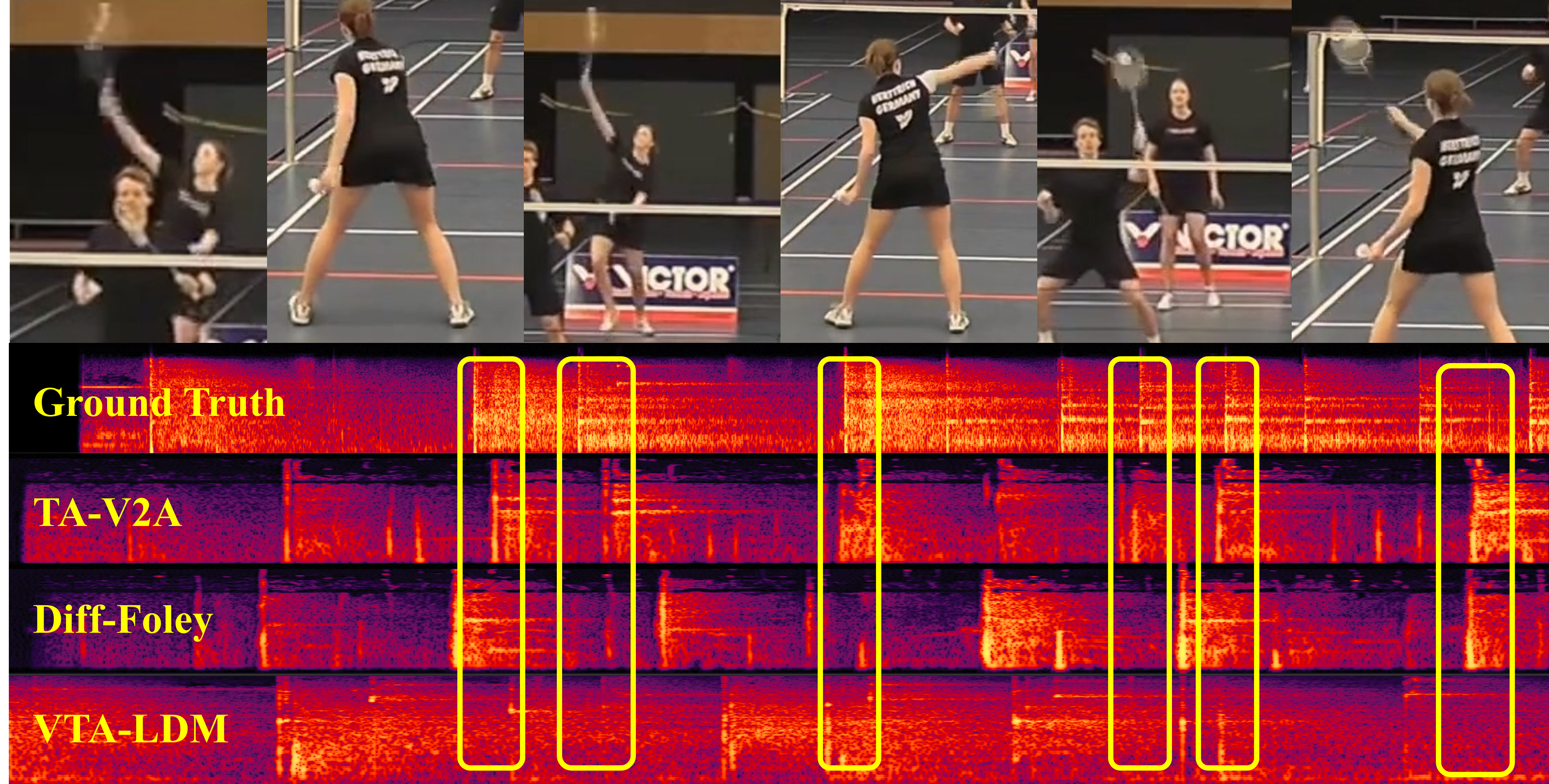}
    }
\caption{An Example of Video-Audio Alignment. The top shows frames from a badminton sequence, while the bottom compares audio spectrograms from different methods: Ground Truth, TA-V2A, Diff-Foley, and VTA-LDM. Yellow boxes highlight key synchronized moments between video and audio.}
\label{fig:align}
\end{figure}

Fig. \ref{fig:align} compares the generated audio from different models, with a focus on our TA-V2A model. The video frames shown at the top depict a badminton match, featuring multiple instances of shuttlecock hits, which are also the key focus for the audio generation models.
 The audio spectrogram generated by our model closely mirrors the ground truth in terms of both timing and frequency. This indicates that the TA-V2A model accurately captures the key audio events, such as the sharp sounds of racket impacts and the shuttlecock's flight, all while maintaining precise temporal alignment with the visual actions in the video.

\noindent \textbf{Subjective Evaluation Results}
Table \ref{tab:tab2} presents the subjective evaluation results of the models. The TA-V2A model, when user modifications were applied during the inference phase, achieved the highest MOS for semantic consistency. This suggests that the integration of a text control interface, along with the PPPR text expansion method, convincingly enhances the model's ability to produce audio that is semantically consistent with the video and closely aligned with human understanding.


Indeed, as evidenced in the analysis above, semantic expression and temporal alignment are intricately connected rather than entirely independent variables. Only with high-quality recognition and generation can the accuracy of alignment be meaningfully evaluated.

\section{Conclusion}
\label{sec:con}

We present TA-V2A, an innovative system for text-assisted video-to-audio generation, featuring a pretraining method that aligns video, text, and audio using advanced diffusion guidance techniques. It includes a text interface for personalized sound generation. Extensive evaluations show that TA-V2A outperforms existing methods in both objective and subjective assessments, enhancing semantic expression. We aim to advance more human-centered, context-aware sound generation. We hope to push the field toward more human-centered, context-aware sound generation.


\vfill\pagebreak

\bibliographystyle{IEEEbib}
\bibliography{refs}

\end{document}